\def\year{2021}\relax
\documentclass[letterpaper]{article} 
\usepackage{aaai21}  
\usepackage{times}  
\usepackage{helvet} 
\usepackage{courier}  
\usepackage[hyphens]{url}  
\usepackage{graphicx} 
\usepackage{natbib}  
\usepackage{caption} 
\title{CogNet: Bridging Linguistic Knowledge, World Knowledge and \\Commonsense Knowledge}

\author{
    Chenhao Wang\textsuperscript{\rm 1,2},
    Yubo Chen\textsuperscript{\rm 1,2},
    Zhipeng Xue\textsuperscript{\rm 1},
    Yang Zhou\textsuperscript{\rm 1,2},
    Jun Zhao\textsuperscript{\rm 1,2}
    \\
}
\affiliations{
    \textsuperscript{\rm 1}National Laboratory of Pattern Recognition, Institute of Automation, Chinese Academy of Sciences, Beijing, China\\
    \textsuperscript{\rm 2}School of Artificial Intelligence, University of Chinese Academy of Sciences, Beijing, China\\
    \{chenhao.wang, yubo.chen, zhipeng.xue, yang.zhou, jzhao\}@nlpr.ia.ac.cn
}

\begin{document}

\maketitle

\begin{abstract}
In this paper, we present CogNet, a knowledge base (KB) dedicated to integrating three types of knowledge: (1) linguistic knowledge from FrameNet, which schematically describes situations, objects and events. (2) world knowledge from YAGO, Freebase, DBpedia and Wikidata, which provides explicit knowledge about specific instances. (3) commonsense knowledge from ConceptNet, which describes implicit general facts. To model these different types of knowledge consistently, we introduce a three-level unified frame-styled representation architecture. To integrate free-form commonsense knowledge with other structured knowledge, we propose a strategy that combines automated labeling and crowdsourced annotation. At present, CogNet integrates 1,000+ semantic frames from linguistic KBs, 20,000,000+ frame instances from world KBs, as well as 90,000+ commonsense assertions from commonsense KBs. All these data can be easily queried and explored on our online platform, and free to download in RDF format for utilization under a CC-BY-SA 4.0 license. The demo and data are available at http://cognet.top/.
\end{abstract}

\section{Introduction}
Over the past decades, knowledge bases (KBs) have shown importance in various artificial intelligent applications. Existing KBs involve roughly three types of knowledge: \textit{linguistic knowledge}, \textit{world knowledge}, and \textit{commonsense knowledge}. To better illustrate, we note some representative projects here:
(1) WordNet \cite{miller_wordnet_1995} and FrameNet \cite{baker_berkeley_1998} provide us with
essential \textbf{linguistic knowledge} respectively from the view of concept relations and frame semantics. For example, one meaning of \texttt{buy} has hyponymy relation to \texttt{get} in WordNet, while in FrameNet \texttt{buy} can evoke a \texttt{Commerce\_buy} frame involving participants like \texttt{Goods} and \texttt{Buyer}. (2) DBpedia \cite{lehmann_dbpedia_2015} and Wikidata \cite{vrandecic_wikidata_2014} mainly focus on explicit \textbf{world knowledge}, containing facts about specific instances, e.g. \texttt{London}$\rightarrow$\texttt{LocatedIn}$\rightarrow$\texttt{England}. (3) ConceptNet \cite{speer_conceptnet_2017} and ATOMIC \cite{sap_atomic_2019} are KBs containing \textbf{commonsense knowledge}.
They try to capture implicit general facts and regular patterns in our daily life, which are loosely organized in free-form expressions, such as ``buy book''$\rightarrow$\texttt{hasPrerequisite}$\rightarrow$``go to bookstore'', where ``buy book'' and ``go to bookstore'' are expressed in natural language phrases, while \texttt{hasPrerequisite} is a pre-defined relation type.

Most of the above KBs focus on a single type of knowledge. However, 
agents with better cognitive intelligence should have the ability to jointly exploit different types of knowledge.
Taking natural language reasoning task as an example, given a sentence
``Emile wants to buy \textit{Hamlet}, so he should go to \underline{\quad} (a book store/a mall/a hospital)'', to choose the most suitable word on the underline, we should know that the verb ``buy'' evokes a frame \texttt{Commerce\_buy} with frame elements \texttt{Buyer} and \texttt{Goods} (linguistic knowledge), and the goods \textit{Hamlet} is a book (world knowledge), as well as ``go to a bookstore'' is a prerequisite for ``buy book'' (commonsense knowledge).
To support such applications, it is an effective method to integrate different types of KBs. 
To this end, there have been several KB projects like YAGO \cite{suchanek_yago:_2007} and FrameBase \cite{gandon_framebase_2015}, both of which built a schema based on linguistic knowledge to organize extensive world knowledge. Another related project Framester \cite{gangemi_framester_2016} integrated several linguistic-factual resources using a formal version of frame semantics.
However, none of them have explicitly integrated linguistic, world and commonsense knowledge altogether. For this target, there are mainly two challenging problems: (1) How to model three different types of knowledge using a unified representation architecture? (2) How to integrate knowledge instances from different KBs, especially integrate the free-form expressions from commonsense KBs with structured tuples from linguistic/world KBs? 

In this paper, we present \textbf{CogNet} (\textbf{Cog}nition \textbf{Net}work), a knowledge base that attempts to bridge linguistic knowledge, world knowledge and commonsense knowledge.
To solve the first problem, we introduce a three-level representation architecture, including semantic frame, frame with element restriction, and frame instance. This is an extension to FrameBase, which built an RDF schema from FrameNet and proposed to represent world knowledge as frame instance. To solve the second problem, we combine automated labeling with crowdsourcing annotations to convert free-form expressions from ConceptNet into structured representations, and further link them with structured knowledge from other KBs (FrameNet, YAGO, Freebase, DBpedia and Wikidata).
At present, CogNet integrates 1,000+ semantic frames from linguistic KBs, 20,000,000+ frame instances from world KBs, as well as 
90,000+ commonsense assertions from commonsense KBs. In addition, we establish an online platform, providing a unified engine to query and explore all these knowledge resources.
 
\begin{figure}
    \centering
    \includegraphics[width=0.85\columnwidth,height=0.25\textheight]{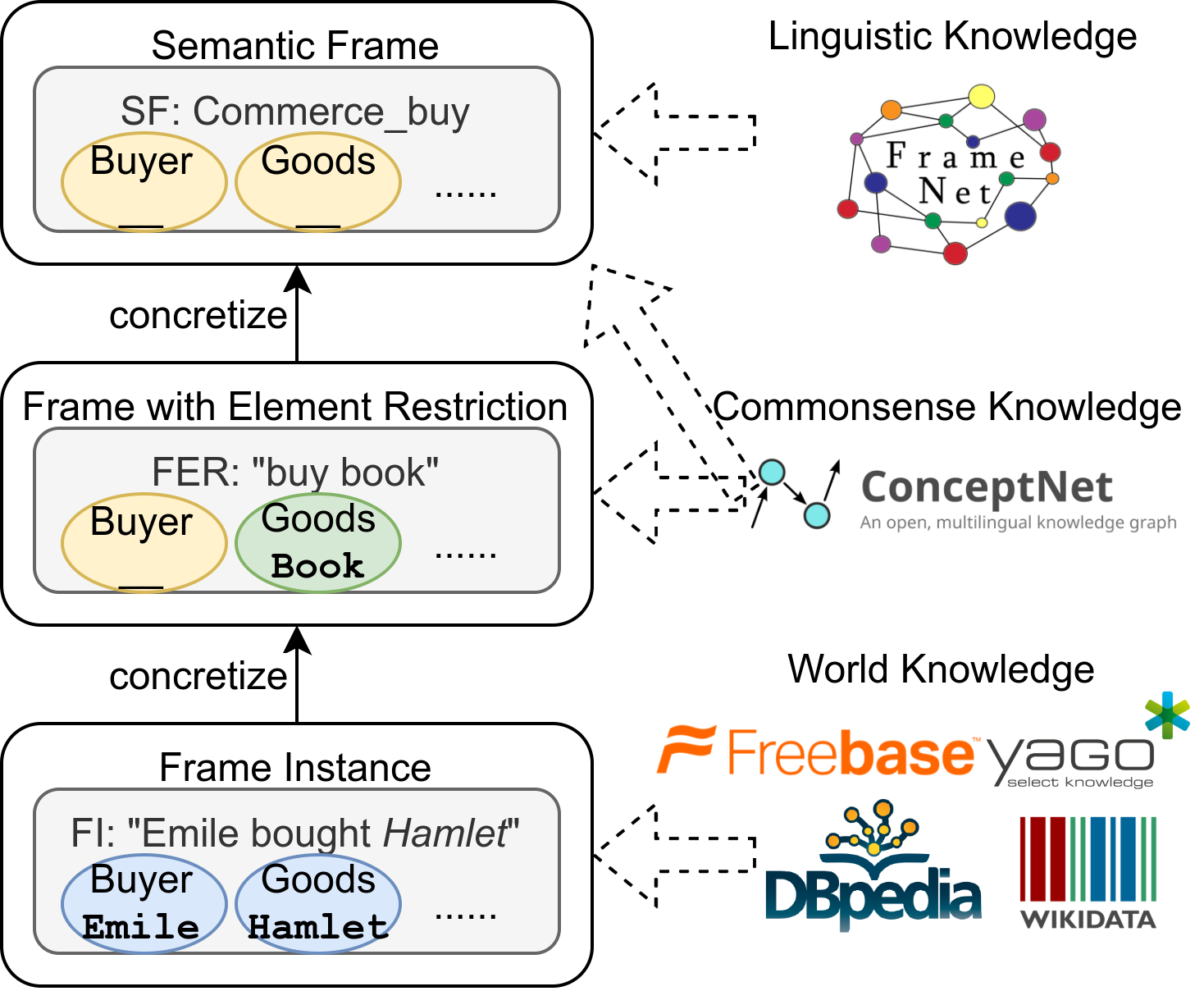} 
    \caption{The 3-level representation architecture of CogNet}
    \label{fig1}
\end{figure}
\section{Methods}

As shown in Figure \ref{fig1}, in CogNet, we model knowledge in three levels of frame-styled representations, namely Semantic Frame (\textbf{SF}), Frame with Element Restriction (\textbf{FER}), and Frame Instance (\textbf{FI}), forming a chain from abstract to concrete: 
\textbf{SF}$\rightarrow$\textbf{FER}$\rightarrow$\textbf{FI}. We design different integrating strategies for each level as follows.

\subsubsection{Semantic Frame} We take SFs and their attached frame elements (FEs) from FrameNet as the core of CogNet, which provides linguistic knowledge about particular situations, objects and events. 
As illustrated in the upper rounded rectangular box in Figure \ref{fig1}, SF \texttt{Commerce\_buy} is associated with FEs like \texttt{Buyer} and \texttt{Goods}. We utilize SFs from FrameBase, which has converted FrameNet into an RDF schema. To extend its coverage, we update it with the latest data from FrameNet 1.7 as well as the Chinese FrameNet \cite{hao_description_2007}.
In addition, we enrich the relations of SFs with commonsense knowledge from ConceptNet. 
\subsubsection{Frame with Element Restriction} We introduce FER nodes
to represent natural language phrases, which extensively exist in free-form commonsense KBs and actually describe frames and their element restriction. For example in ``buy book''$\rightarrow$ \texttt{hasPrerequisite}$\rightarrow$``go to bookstore'', ``buy book'' tells us the type of \texttt{Goods}. 
As illustrated in the middle rounded rectangular box in Figure \ref{fig1}, we convert the phrase into a FER, which concretizes SF \texttt{Commerce\_buy} and is associated with a restriction that the FE \texttt{Goods} belongs to \texttt{Book} type.
We extract FERs from ConceptNet via three steps: (1) parsing the concepts into frames with their elements through automated and crowdsourced methods; (2) disambiguating the FEs in the last step and linking them to WordNet taxonomy; (3) connecting FERs generated in previous steps with commonsense relations. Note that a concept from ConceptNet could be mapped to either a FER or a frame, depending on its content. That is why there are two arrows from ConceptNet in Figure \ref{fig1}.
\subsubsection{Frame Instance} 
We represent world knowledge facts from different sources as FIs. 
In a FI, FEs are bound with specific objects or literals. For example, as shown in the lower rounded rectangular box in Figure \ref{fig1}, ``Emile bought \textit{Hamlet}'' is corresponding to a FI where \texttt{Buyer} is \texttt{Emile} and \texttt{Goods} is \texttt{Hamlet}. We extract 
FIs from YAGO, Freebase, DBpedia, and Wikidata based on SPARQL rules of FrameBase, and merge the same entities from different sources. Further, we connect FIs with FERs through ``concretize'' relations if there are just some FERs describing the corresponding general situation. 
\begin{figure}
	\centering
	\includegraphics[width=1\columnwidth,height=0.25\textheight]{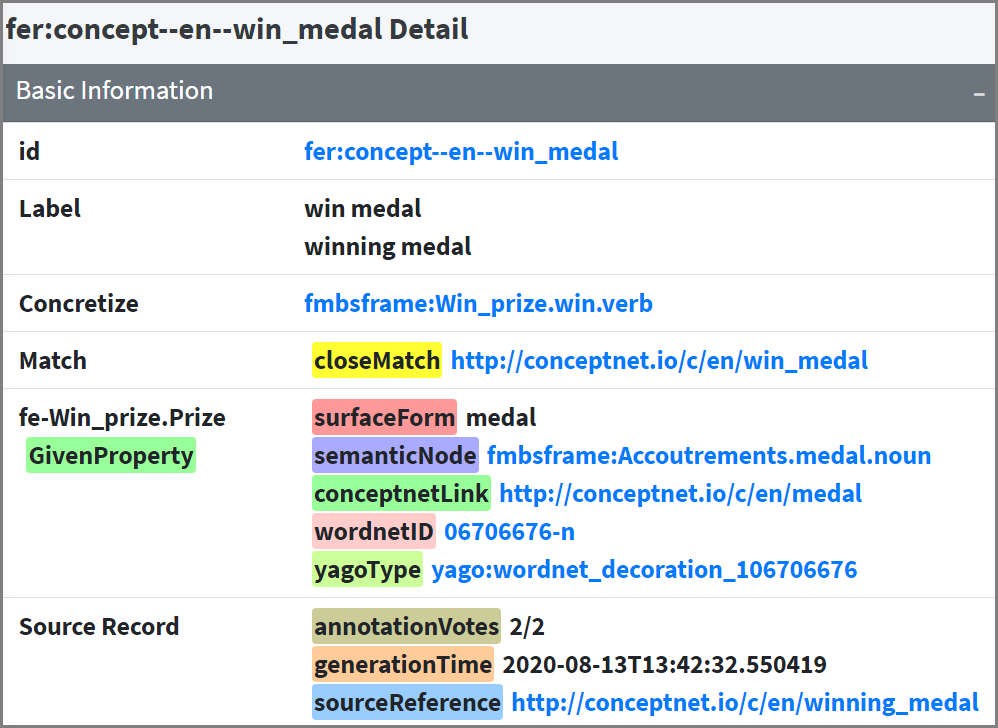} 
	\caption{An example of exploring page.}
	\label{fig2}
\end{figure}

\section{Online Platform}
We develop an online platform for access to CogNet data, including key-word querying and top-down exploring.

\subsubsection{Querying}
We provide an interface for unified queries of different types of knowledge, which can return a list of candidate nodes according to the query. Specifically, the system can detect frame names or lexical units in the query and find the corresponding semantic frames. Then, it can find FERs and entities involved in FIs by fuzz-matching the query with their labels. Finally, the system can aggregate the results and return them to users. Besides the easy-using interface for ordinary users, we provide SPARQL query service to skilled users so that they can implement more complex queries.

\subsubsection{Exploring} 
As CogNet is organized with different levels of frame representations, we provide a top catalog for semantic frames, from which users can explore knowledge from the top down via ``concretize'' relations. For each SF, FER, FI, and entity involved in FIs, we provide a page to show its description and relationship to other nodes. A typical page is shown in Figure \ref{fig2}. We display the label, frame elements, reference sources, relevant nodes, and other useful information on the page. For other nodes, the displayed information may vary according to the type. Refer to http://cognet.top/ for more details.

\section{Acknowledgments}
This work is supported by the National Key R\&D Program of China (2020AAA0106400), the National Natural Science Foundation of China (No.61533018, No.61976211, No.61806201). This work is also supported by the independent research project of National Laboratory of Pattern Recognition. We thank ConceptNet \cite{speer_conceptnet_2017}, DBpedia \cite{lehmann_dbpedia_2015}, FrameBase \cite{gandon_framebase_2015}, FrameNet \cite{baker_berkeley_1998}, Framester \cite{gangemi_framester_2016}, Freebase \cite{bollacker_freebase:_2008}, Predicate Matrix \cite{de_lacalle_predicate_2014}, PreMOn \cite{corcoglioniti_premon_2016}, Semlink \cite{bonial_renewing_2013}, Wikidata \cite{vrandecic_wikidata_2014}, WordNet \cite{miller_wordnet_1995} and YAGO \cite{suchanek_yago:_2007} for their previous contributions to creating and integrating different types of open knowledge resources. We also thank the CFN Research Team of Shanxi University for providing the Chinese FrameNet resources.

\bibliography{mylib.bib}
\end{document}